\documentclass[10pt,twocolumn,letterpaper]{article}

\usepackage{cvpr}              

\usepackage{times}
\usepackage{epsfig}
\usepackage{graphicx}
\usepackage{amsmath}
\usepackage{amssymb}
\usepackage{mathtools}
\usepackage{dsfont}
\usepackage[mathscr]{euscript}
\DeclareMathOperator*{\argmin}{arg\,min}
\newcommand\norm[1]{\lVert#1\rVert}
\usepackage{color, colortbl,xcolor}
\definecolor{Gray}{gray}{0.9}
\definecolor{light-gray}{gray}{0.95}

\usepackage{pifont}
\newcommand{\cmark}{\ding{51}}%
\newcommand{\xmark}{\ding{55}}%

\usepackage{booktabs} 
\usepackage{multirow} 
\usepackage{subcaption} 
\usepackage{url} 

\usepackage{microtype}
\usepackage{cuted} 
\usepackage{capt-of}
\usepackage[font={small}]{caption}
\usepackage{epigraph}
\usepackage{comment}
\newcommand\sbullet[1][.5]{\mathbin{\vcenter{\hbox{\scalebox{#1}{$\bullet$}}}}}

\usepackage[pagebackref,breaklinks,colorlinks]{hyperref}

\usepackage[capitalize]{cleveref}
\crefname{section}{Sec.}{Secs.}
\Crefname{section}{Section}{Sections}
\Crefname{table}{Table}{Tables}
\crefname{table}{Tab.}{Tabs.}

\def\cvprPaperID{812} 
\def\confName{CVPR}
\def\confYear{2022}

\begin{document}

\title{Learning to Listen: Modeling Non-Deterministic Dyadic Facial Motion}

\author{Evonne Ng\textsuperscript{1}
\hspace{0.3in} Hanbyul Joo\textsuperscript{2}
\hspace{0.3in} Liwen Hu\textsuperscript{3}
\hspace{0.3in} Hao Li\textsuperscript{3}
\hspace{0.3in} Trevor Darrell\textsuperscript{1}\\
\hspace{0.3in} Angjoo Kanazawa\textsuperscript{1}
\hspace{0.3in} Shiry Ginosar\textsuperscript{1}
\vspace{10pt}
\\
\textsuperscript{1}{UC Berkeley}
\hspace{0.3in} \textsuperscript{2}{Seoul National University} 
\hspace{0.3in} \textsuperscript{3}{Pinscreen} \\ 
}

\maketitle


\begin{strip}\centering
\includegraphics[width=\linewidth]{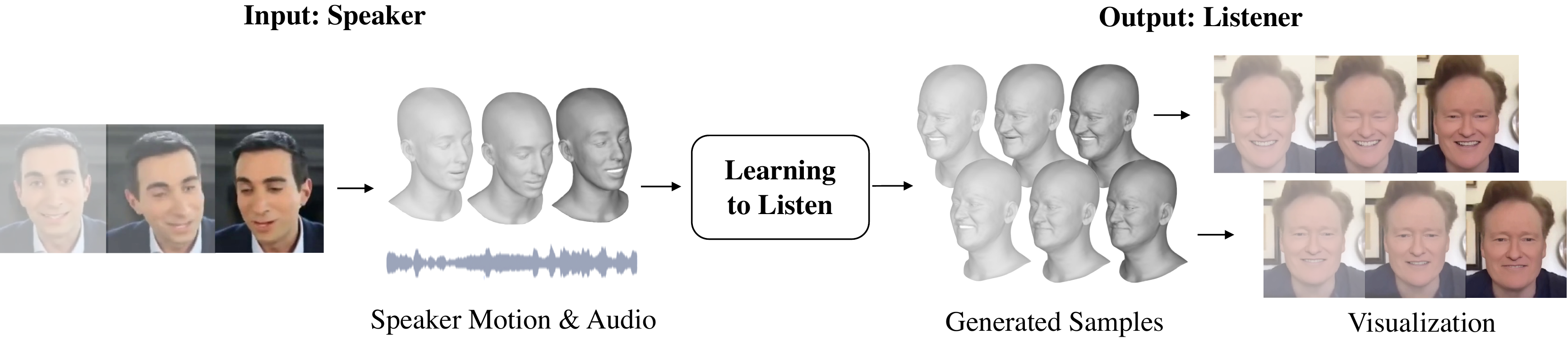}
\captionof{figure}{{\bf Synthesizing listeners.}
Given a speaker video, we extract the audio and motion of the speaker. From these multimodal speaker inputs, our method 
synthesizes multiple realistic listener 3D motion sequences (top and bottom) in an autoregessive fashion. 
The output of our approach can be optionally rendered as photorealistic video. Please see \href{https://youtu.be/3CSlKZ5T6DU}{results video}.
}
\label{fig:teaser}
\end{strip}

\begin{abstract}
We present a framework for modeling interactional communication in dyadic conversations: given multimodal inputs of a speaker, we autoregressively output multiple possibilities of corresponding listener motion. We combine the motion and speech audio of the speaker using a motion-audio cross attention transformer. Furthermore, we enable non-deterministic prediction by learning a discrete latent representation of realistic listener motion with a novel motion-encoding VQ-VAE. Our method organically captures the multimodal and non-deterministic nature of nonverbal dyadic interactions. Moreover, it produces realistic 3D listener facial motion synchronous with the speaker (see video). We demonstrate that our method outperforms baselines qualitatively and quantitatively via a rich suite of experiments. To facilitate this line of research, we introduce a novel and large in-the-wild dataset of dyadic conversations. Code, data, and videos available on \href{https://evonneng.github.io/learning2listen/}{project page}.

\end{abstract}

\section{Introduction}
\label{sec:intro}
\vspace{-.05in}
\epigraph{\textit{``Thus the body of the speaker dances in time with his speech. Further, the body of the listener dances in rhythm with that of the speaker!"}}{\sc{--- Condon and Ogston,} \em{1966}}
\vspace{-.05in}

When we speak, it is rarely in a void --- rather, there is often a listener at the other end of the conversation.
As a speaker, we are acutely aware of what the listener is doing. A slight off-sync motion or a diverted look may throw us off, suggesting the listener is bored or otherwise preoccupied, leaving us feeling misunderstood~\cite{kendon1970movement}. Indeed, successful conversations rely on a coordinated dance between the speaker and the listener in which the two signal to each other that they are communicating with one another and not with anyone else~\cite{kendon1970movement}. This chameleon effect~\cite{chartrand1999chameleon} of nonverbal mimicry during conversation results in smoother interactions, increases the liking between interaction partners, establishes rapport~\cite{LaFrance1979}, and may even predict the long term outcome of psychotherapy~\cite{psychotherapy_2014}.
Interestingly, nonverbal feedback from a listener, such as head movement is more central to keeping a conversation flowing than
content-based replies~\cite{CassellThorisson1999}.
In this work, we propose a computational framework that can similarly provide nonverbal feedback in response to a speaker in a contextual and timely manner. Such an ability is critical for virtual agents to meaningfully interact with humans, for whom nonverbal communication is central
from infancy~\cite{infant_itersubjectivity_1979}. 

Modeling nonverbal feedback during dyadic interaction is a difficult problem, as listener responses are nondeterministic in nature. Moreover, speakers are inherently multimodal, as they communicate both verbally via speech, and nonverbally via face and body motion. Capturing interaction in its natural setting requires addressing both challenges.
The task of modeling human conversations has a long history.
However,
unlike traditional rule-based methods~\cite{Cassell1994,gratch2006virtual,Bohus_Horvitz_2010,huang2011virtual} or methods that rely on modeling hand-defined simple motion characteristics such as smiles~\cite{movellan2015} or head nods~\cite{gratch2006virtual,huang2011virtual}, we wish to model the true complexity of the interaction.
This is hard to achieve and generalize using conventional database methods
that generate motion via a lookup into a database of ground truth motion~\cite{zhang2008spacetime,jung2011believable,serra2016behavioural}.
We, therefore, learn to model these 
dyadic conversational dynamics implicitly in a data-driven way by directly observing human conversations in in-the-wild videos.

Given a video of a speaker, we extract their speech audio, and facial motion (Figure~\ref{fig:teaser}(left)). We combine information from both modalities using a motion-audio cross-attention transformer. From this multimodal speaker input, we learn to autoregressively synthesize multiple modes of motion representing different possible responses of a listener who moves 
synchronously with
the speaker (Figure~\ref{fig:teaser}(right)).

Modeling the nondeterminism in listener responses is a key element in capturing conversational dynamics. Previous attempts to tackle this problem applied various techniques but fell short of achieving realistic outputs~\cite{jonell2020let}. We propose
to learn a realistic manifold of listener motion by quantizing the space of listener motion with a novel sequence-encoding VQ-VAE~\cite{van2017neural}, which efficiently captures a wide range of motion in a discrete format that is well-suited for learning. To the best of our knowledge, we are the first to extend VQ-VAE models to the domain of motion synthesis. The learned discrete codebook of listener motion allows us to predict a multinomial distribution of future motion. From this distribution we can sample a wide range of possible modes of motion representing different perceptually-plausible listeners, capturing their inherent non-deterministic nature. Furthermore, we demonstrate our learned discrete latent codes can stay on the manifold of realistic motion ensuring no motion drift occurs even in long-horizon predictions.
Meanwhile, the autoregressive nature of our method allows us to consider speaker sequences of any length.

To support our data-driven approach to modeling human conversation, 
data is needed in the form of videotaped dyadic interactions where both parties are ideally filmed from a head-on frontal view.
This kind of data is hard to come by.
While the first investigation of interactional synchrony in conversation
dates back to Condon and Ogston in 1966~\cite{condon_ogston_1966}, current studies
still mostly rely on in-lab footage~\cite{huang2011virtual,Hart2016,gaziv2017reduced,cheong_molani_sadhukha_chang_2020} or small-scale motion-capture datasets~\cite{busso2008iemocap,jonell2020let}.
Notable exceptions are~\cite{learn2smile2017,nojavanasghari2018interactive}, yet the footage has not been made publicly available.
We collect a large-scale source of data in the form of split-screen recorded online interviews where the speaker and listener are captured in frontal view. Our dataset, which consists of 72 hours of in-the-wild conversations, enables the investigation of dyadic communication using the latest machine learning methods.

We evaluate the synthesized listener motion compared to ground truth as well as baseline methods and ablations via an extensive quantitative study. We employ a wide array of metrics to test the realism and diversity of the synthesized motion,
and the synchronization of the listener's motion with that of the speaker. While measuring realism and diversity centers on the generated motion of the listener in isolation, synchrony captures aspects of the dyad as a whole. We further corroborate our qualitative findings by inviting human observers to evaluate our results. While we assess our method using the raw 3D mesh output, we additionally illustrate our results by translating the 3D output to pixels for viewing purposes only, as synthesized video provides a richer perceptual context. Under both quantitative and qualitative measures, our method significantly outperforms all baselines. Our synthesized listeners were deemed plausible by human observers when compared to ground-truth motion. 
This highlights our method's ability to produce realistic-looking motion that is synchronous with a given speaker.

Our main contribution is in our learning-based approach towards understanding human interactional communication in conversation. 
We combine multimodal speaker inputs via motion-audio cross-attention. We extend vector quantization  
to the domain of motion synthesis and learn a quantized space of motion in which we autoregressively predict multiple modes of perceptually realistic listener motion.
To support future endeavors in this direction we \textit{publicly release} a novel dataset of 72 hours of in-the-wild dyadic conversational videos with detailed 3D annotations capturing subtleties in expression and fine-grain head motion.

\section{Related Work}
\label{sec:relatedworks}

We discuss related works concerned with conversational agents and motion synthesis. For a review of interactional motion in human communication, see Appendix~\ref{appendix:rw}.

\smallskip
\noindent \textbf{Interactional Motion in Conversational Agents.}
Prior works on conversational avatars manually incorporated different aspects of interactional motion~\cite{Cassell1994, gratch2006virtual, huang2011virtual, Bohus_Horvitz_2010, sonlu2021conversational}. These approaches designed rule-based methods to generate agents that can interact via appropriate facial
gestures \cite{gratch2006virtual, huang2011virtual, sonlu2021conversational}, speech \cite{Cassell1994}, or a combination of modalities~\cite{Bohus_Horvitz_2010}. All these methods use lab-recorded motion capture sequences.
These either limit the variety of captured gestures, or rely on simplifying assumptions for motion generation which do not hold for in-the-wild data.

Prior data-driven methods predict the 2D motion of one person in a conversation as a function of the other's motion~\cite{learn2smile2017,nojavanasghari2018interactive}. These require a pre-defined dictionary achieved by clustering 
motion frequencies or 2D facial keypoints from the training set. Rather, we reason in 3D and learn a discretized latent space that captures the manifold of facial motion. Other methods using 3D investigate interactional dynamics while focusing on full 3D body motion and turn taking~\cite{joo2019towards, ahuja2019react}. Others tackle the problem of facial gestures in conversation by simplifying the task to predicting head nods~\cite{ahuja2019react}, estimating head pose~\cite{greenwood2017predicting}, or generating a single image of a facial expression that summarizes the entire speaker 
sequence~\cite{huang2017dyadgan,nojavanasghari2018interactive}. In contrast, our method captures the natural complexity of interactions by considering the full range of facial expressions and head rotations.

Recent methods began generating 3D facial motion with additional inputs from the listener such as text~\cite{chu2018face} or speech~\cite{jonell2019learning,jonell2020let}. Most similar to our approach is that of Jonel~\etal~\cite{jonell2020let}, who propose a Glow-based method~\cite{kingma2018glow,henter2020moglow}. However,
 their method takes as input the full temporal context of listener audio and is reported to perform better without any audio input. 
In contrast, our method does not use any listener audio as additional input.
Additionally, we quantitatively demonstrate that \textit{each of the input modalities is essential} to its performance.

\smallskip
\noindent \textbf{Conditional Motion Synthesis.} 
Gestural motion synthesis has previously relied on convolutional auto-encoders to learn a representation of human motion~\cite{ng2021body2hands, ginosar2019learning, learn2smile2017, joo2019towards, jonell2020let}.
Some methods incorporated an adversarial loss~\cite{ng2021body2hands, ginosar2019learning} or experimented with flow models~\cite{jonell2020let} and other sampling-based methods~\cite{learn2smile2017} to generate more diverse and realistic motion. Recent works  demonstrated the success of using transformers in generating diverse motion with long-range dependencies~\cite{petrovich2021action, cao2020long, li2021learn, li2020learning}.
These generate possible motion segments conditioned on action~\cite{petrovich2021action}, 3D human motion trajectory in a scene conditioned on a goal~\cite{cao2020long}, or dance motion from audio~\cite{li2020learning,li2021learn}. Similarly, we employ a transformer-based predictor for conditional motion synthesis.
Additionally, to the best of our knowledge, we are the \emph{first to demonstrate the benefits of using vector quantization (VQ-VAE~\cite{van2017neural}) to achieve improved motion synthesis results}. In essence, rather than relying on the addition of Perlin noise~\cite{perlin1996improv} for improved realism, we learn the fine details of realistic motion in a data-driven way.

\section{Method}
\label{sec:method}
\begin{figure}[t]
\begin{center}
\includegraphics[width=\linewidth]{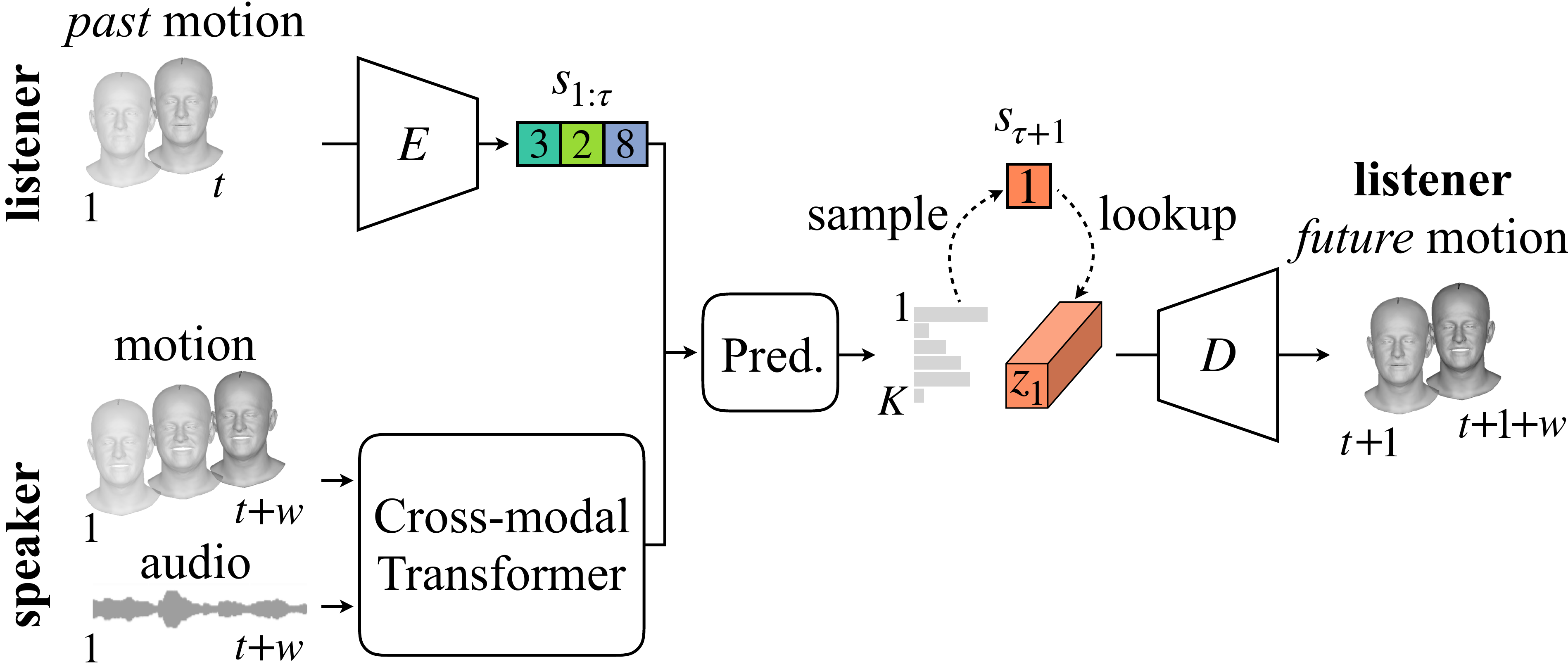}
\end{center}
\vspace{-1em}
  \caption{\textbf{Overview}: We predict a distribution over future listener motion conditioned on multimodal inputs from a speaker. We use cross-modal attention to fuse the speaker audio and motion input, and a novel sequence-encoding VQ-VAE to discretize past listener motion. Our autoregressive Predictor
  outputs a distribution over the $K$ discrete codebook indices, from which we sample a code for the next timestep. We obtain the continuous future listener motion by decoding the sampled codebook index.}
\label{fig:method}
\end{figure}

Our goal is to model the conversational dynamics between a speaker and a listener.
To test whether our model captures the subtleties of face-to-face communication, 
we synthesize the interactional motion responses of the listener, which are known to be essential to the flow of conversation~\cite{kendon1970movement,LaFrance1979,chartrand1999chameleon}. 

We define the following task:
\emph{given the 3D facial motion and audio of the speaker, we autoregressively predict the corresponding facial motion of the listener.}

To represent the ongoing flow of conversation, we define a transformer-based predictor, $\mathcal{P}$, that learns to model temporally long-range patterns in the input sequence (Sec.~\ref{sec:predictor}).
The predictor takes two inputs: one corresponding to the speaker and the other to the listener (Figure~\ref{fig:method}).

To model the speaker's audio and facial motion, we introduce a motion-audio cross-modal transformer that learns to fuse the two modalities (Sec.~\ref{sec:crossmodal}).
To represent the manifold of realistic listener facial motion, we extend VQ-VAE~\cite{van2017neural} to the domain of motion synthesis and learn a codebook of a discrete latent space  (Sec.~\ref{sec:codebook}).
This discrete representation enables us to predict a multinomial distribution over the next timestep of motion. Thus,
the output of the autoregressive predictor is a distribution over possible synchronous
and realistic 
listener responses, from which we can sample multiple trajectories.

\subsection{Problem Definition}\label{sec:motion_rep}
Let $\mathbf{F}=\{\mathbf{f}_i\}_{i=1}^T$ be a temporal sequence of facial motions $\mathbf{f}_i$. 
We use $\mathbf{F}^S$ and $\mathbf{F}^L$ to denote the motion of the speaker and listener respectively.
For each timestep $t \in [1,T]$, we take as input a speaker's facial motion 
$\mathbf{F}^S_{1:t}=(\mathbf{f}^S_{1}, \cdots, \mathbf{f}^S_{t})$
and their corresponding speaker audio sequence $\mathbf{A}^S_{1:t}$, 
along with any previously predicted past listener motion $\mathbf{\hat{F}}^L_{1:t-1}$, 
if available. 
Our predictor, $\mathcal{P}$, then autoregressively predicts the corresponding listener facial motion one time-step at a time:
\begin{equation}
       \mathbf{\hat{f}}^L_{t} = \mathcal{P}(\mathbf{F}^S_{1:t}, \mathbf{A}^S_{1:t}, \mathbf{\hat{F}}^L_{1:t-1}), \\
    \label{eq:motion_eq}
\end{equation}
where $\mathcal{P}$ learns to model the distribution over the next timestep of listener motion
\begin{equation}
p(\mathbf{\hat{f}}^L_{t} | \mathbf{F}^S_{1:t}, \mathbf{A}^S_{1:t}, \mathbf{F}^L_{1:t-1}).
\end{equation}

To obtain speaker-only audio, we filter out all listener audio back-channels using sound source separation~\cite{owens2018audio}. To represent the motion, we estimate the 3D facial expressions and orientations from video frames of human conversations using a 3D Morphable Face Model (3DMM)~\cite{blanz1999morphable,paysan20093d,cao2013facewarehouse, FLAME:SiggraphAsia2017}. 3DMMs are parametric facial models that allow us to directly regress disentangled coefficients corresponding to facial expression, head orientation, and identity-specific shape from a single image~\cite{zollhofer2018state}. This process results in facial expression coefficients $\beta_t \in \mathds{R}^{d_m}$, where $d_m$ is the dimension of the expression coefficient, a normalized 3D head pose $R_t \in SO(3)$, and shape coefficents that we discard to obtain an identity-agnostic representation. 
Our facial representation at time $t$, $\mathbf{f}_t \in \mathds{R}^{d_m+3}$, is a concatenation of expression and orientation (in Euler angles):
\begin{equation}
    \mathbf{f}_t = [\beta_t,  R_t].
\end{equation}
We normalize facial orientation by computing the mean frontal face direction per video (\ie, orientation at rest pose) and align all head poses in the sequence with respect to this rest pose. This allows us to achieve a camera-view agnostic representation.
In contrast to the 2D representations used in some prior works~\cite{learn2smile2017,nojavanasghari2018interactive}, our 3D representation is invariant to changes in facial shape, scale, and camera pose, allowing us to generalize across new faces and camera viewpoints.


\subsection{Quantized Listener Motion Codebooks}
\label{sec:codebook}

\begin{figure*}
\begin{center}
\includegraphics[width=\linewidth]{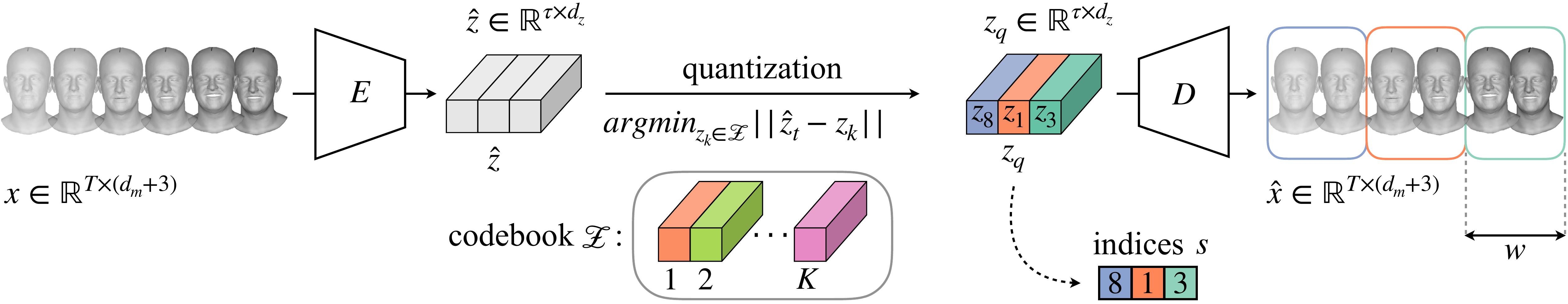}
\end{center}
    \vspace{-1.5em}
  \caption{\textbf{Motion VQ-VAE that learns a discrete listener motion codebook.} The input is a $T$ length sequence of raw listener facial motion (expression coefficients and 3D head rotations). The \textit{transformer} sequence-encoder $E$ compresses the input into an embedding that gets mapped to its closest quantized codebook element in $\mathcal{Z}$. The \textit{transformer} decoder $D$ decodes the quantized embedding into an approximate reconstruction of the input. We train on a reconstruction loss and commitment loss (Eq.~\ref{eq:vqloss}). Not only does the VQ-VAE allow us to learn a representation robust to drift from autoregressive inference,
  it also enables non-deterministic motion synthesis.
  }
\label{fig:vq}
\end{figure*}

We extend the use of VQ-VAE~\cite{van2017neural} to produce multiple realistic modes of different listener responses. VQ-VAE was originally proposed as a method to learn a quantized codebook of image elements from which images could be synthesized autoregressively. Convolutional architectures were used both for learning the codebook and for recombining the discrete elements into images~\cite{van2017neural}. While the synthesis step was later replaced by transformer architectures that can learn long-range connections~\cite{esser2021taming}, image-generation approaches employ a convolutional encoder-decoder pair. This is well-suited for images but not for temporal sequences where convolving over the temporal domain may lose high-frequency information. We design a novel sequence-encoding VQ-VAE where we utilize transformers for the encoder-decoder pair.
To the best of our knowledge, we are the first to apply a VQ-VAE to the domain of motion generation.  

The advantages of this method are three-fold: (1) it allows us to predict a  multinomial distribution over future motion from which we can sample many possible output modes, (2) using the learned discrete latent codes allows us to stay on the manifold of realistic motion ensuring no drift occurs (a problem for methods that directly regress continuous outputs~\cite{aksan2019structured}), and (3) it produces realistic motion that captures high-frequency movements.

Specifically, we train a VQ-VAE transformer encoder ${E}$ and decoder ${D}$. 
To handle the temporal nature of the input, we learn to model longer listener motion sequences in terms of shorter temporal components. Rather than considering static expressions/rotations independently, the latent embedding covers multiple frames, allowing it to learn motion dynamics.
The latent embedding represents motion segments of temporal window size $w \ll T$ from a  discrete codebook $\mathcal{Z} = \{z_k\}^{K}_{k=1}$, where $z_k \in \mathds{R}^{d_z}$, that we jointly learn with $E$ and $D$. 
$\mathcal{Z}$ maps each of the $K$ codebook entries to a discrete code element of dimension $d_z$. 
As shown in Figure~\ref{fig:vq}, we can then approximate any raw listener motion segment $x = \mathbf{F}^L_{1:T} \in \mathds{R}^{T \times (d_m+3)}$ of length $T$ 
in three steps. First, we encode the sequence $\hat{z} = E(x) \in \mathds{R}^{\tau \times d_z}$, where $\tau = \frac{T}{w}$ is the length of the patch-wise encoded sequence. 
Second, we obtain the quantized sequence, $z_q$, via an element-wise quantization function $\mathbf{q(\cdot)}$ that maps each element of the encoded sequence $\hat{z}$ to its closest codebook entry:
\begin{equation}
    \label{eq:quant}
    z_q = \mathbf{q}(\hat{z}) \coloneqq \Big( \argmin_{z_k \in \mathcal{Z}}  \norm{\hat{z_t} - z_k} \Big) \in \mathds{R}^{\tau \times d_z}.
\end{equation}
Finally, the reconstruction $\hat{x} \approx x$ is given by: 
\begin{equation}
    \label{eq:recon}
    \hat{x} = D(z_q) = D(\mathbf{q}(E(x))).
\end{equation}
We train $E,D$ and the codebook 
with the loss function~\cite{van2017neural},
\begin{equation}
\label{eq:vqloss}
     \begin{aligned}
        \mathscr{L}_{\text{VQ}}(E,D,\mathcal{Z}) &= \norm{x - \hat{x}}  \\
        & + \norm{\text{sg}[E(x)] - z_q} \\
        & + \norm{\text{sg}[z_q] - E(x)},
    \end{aligned}
\end{equation}
where $\norm{x - \hat{x}}^2$ is a reconstruction loss, sg$[\cdot]$ is a stop-gradient operation, and $\norm{\text{sg}[z_q] - E(x)}^2_2$
is a ``commitment loss"~\cite{van2017neural}. After learning the codebook of listener motion, we use the pretrained encoder to quantize the listener motion input to the predictor (Figure~\ref{fig:method}).


\subsection{Cross-Modal Attention for Speaker Input}
\label{sec:crossmodal}

From the speaker, we take as input both audio $a = \mathbf{A}^S_{1:t+w}$, and facial motion $m = \mathbf{F}^S_{1:t+w}$. Here, $w$ is the amount of additional future context we see from the speaker. This context acts as a feedback delay that is beneficial in improving learned synchrony for robotics~\cite{washburn2019feedback}. In contrast to the listener motion, we do not quantize the speaker inputs. While we experimented with both options, we found that speaker motion quantization did not improve results, and quantizing the audio deteriorated the results significantly. We conclude that while quantization is beneficial for predicted motion, for the quality of results as well as sampling capabilities, it is not advantageous for input modalities.

We learn to fuse the audio and motion modalities together using
cross-modal attention. Cross-model attention of text and audio~\cite{ahuja2020no} or language and vision~\cite{lu2019vilbert,tsai2019multimodal,jaegle2021perceiver} has been shown to outperform early or late fusion. We extend its use to successfully fuse information from \textit{motion} and audio, a task that proved difficult to previous approaches~\cite{jonell2020let}.  We additionally
experimented with a naive method of 
concatenating audio and motion, but 
this resulted in empirically worse results due to overly-long conditioning sequences.
Applying cross-modal attention along a temporal sequence also allows different modalities to discover some temporal re-alignment~\cite{ahuja2020no}. This is especially helpful for encoding speaker inputs since a speaker's motion may not always align with their speech (\eg delay for dramatic effect).

We compute the Queries $Q_a$ for the cross-modal attention operation from the audio input, and the Keys $K_m$ and Values $V_m$ from the motion. 
We then apply a series of cross-modal attention blocks on the motion modality, where the audio queries are always computed from the raw audio:
\begin{equation}
    \text{attention}_{m \rightarrow a} = \text{softmax}\Big( \frac{Q_a K_m^\top}{\sqrt{d_k}}\Big) V_m.
\end{equation}
Here, $d_k$ is the transformer hidden dimension. The cross-modal transformer outputs an intermediate embedding that incorporates information from both the audio and motion of the speaker. Additional convolutional layers temporally downsample the sequence to match the size of the quantized listener sequence.
The final speaker encoding is an embedding $m' \in \mathds{R}^{(\tau+1) \times d_k}$.
We experimentally verify that this method of fusion outperforms others (Table.~\ref{tab:baselines}). 


\subsection{Listener Motion Predictor}
\label{sec:predictor}

We design a transformer-based predictor module, $\mathcal{P}$, to capture long-range correlations in the input data. Building off~\cite{li2021learn}, 
we employ full-attention masking on the inputs, which has shown promising results in generating long-range motion in an auto-regressive manner.  However, with our discrete latent code representation, our model is additionally able to capture multiple modes of outputs by predicting the distribution of possible next motions. Furthermore, we enable multi-modal inputs by means of cross-attention.

\begin{figure*}[t]
\begin{center}
\includegraphics[width=\linewidth]{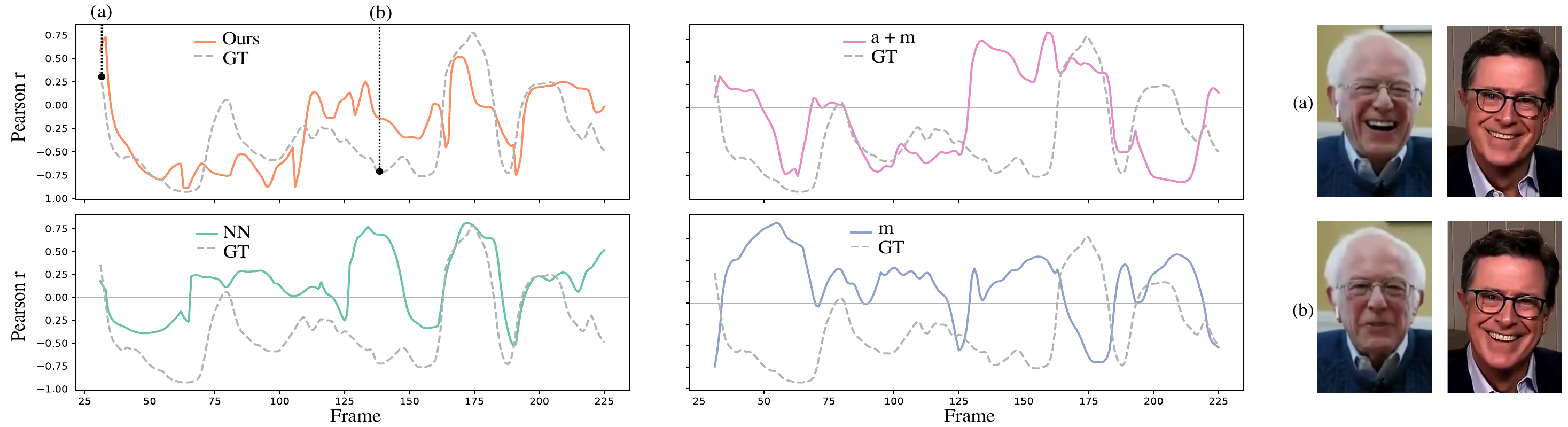}
\end{center}
    \vspace{-1.5em}
   \caption{Synchrony of expressions between speaker and listener measured by PCC across a sequence. We convert the expression sequence to a 1D lip curvature time-series according to~\cite{ginosar2015century}. \textbf{Ours} best matches the synchrony seen in ground truth. \textbf{NN} produces sequences that are too synchronous with the speaker. \textbf{a+m} and \textbf{m} fail to follow the major trends seen in ground truth, such as periods of (a) high synchrony when both the listener and speaker are laughing, and (b) low/no synchrony when the speaker speaks and the listener continues smiling.
   }
\label{fig:sync}
\end{figure*}

$\mathcal{P}$ takes as input
the multimodal speaker embedding $m'$ as well as the sequence of previously predicted listener motion.
Rather than representing the listener quantized motion 
as a sequence of 
codebook vectors $z_q$, for the purpose of
prediction
we use the 
parallel representation of a sequence of
 corresponding codebook \textit{indices}, $s = s_{1:\tau} \in \{1,...,K\}^\tau$. 
Specifically, we discretize past continuous listener motion $x = \mathbf{F}^L_{1:t}$ by encoding it via the pre-trained encoder $E$ and quantization $\mathbf{q}$ (Section~\ref{sec:codebook}). We then obtain the sequence of \textit{indices} of the nearest codebook entry per element, via $\mathbf{I}(\cdot)$, an element-wise inverse-lookup function that returns the index of a given codebook element 
\begin{equation}
\label{eq:index}
   \quad s_{1:\tau} = \textbf{I}(\mathbf{q}(E(x))).
\end{equation}

Given speaker input $m'$ and listener input $s_{1:\tau}$, the predictor outputs $p(s_{\tau+1}) \in \mathds{R}^{K}$, the multinomial distribution of the next listener codebook index across the $K$ entries:
\begin{equation}
    p(s_{\tau+1}) = \mathcal{P}(m', s_{1:\tau}).
\end{equation}

We can then sample from $p(s_{\tau+1})$ to obtain an index $k$ into the codebook $\mathcal{Z}$. We perform a codebook lookup
to retrieve the corresponding quantized element $z_k$ of listener motion, which we pass through the decoder $D$. The output is the predicted continuous future listener motion $\hat{y} = \mathbf{\hat{F}}^L_{t+1:t+1+w}$ of length $w$. We train our network with a cross entropy loss on the codebook index $s_{\tau+1}$:
\begin{equation}
\label{eq:transformerloss}
    \mathscr{L_\mathcal{P}} = \mathds{E}_{y \sim p(y)}[-\log(p(s_{\tau+1})],
\end{equation}
where the target codebook index at $\tau+1$ is computed from ground truth future facial motion $y = \mathbf{F}^L_{t+1:t+1+w}$. 

At train time, we follow teacher-forcing and use ground truth listener motion $y$ as past listener input. We randomly mask prior timesteps $\in [1,\tau]$
to facilitate autoregressive learning. At test time we input zeros for timesteps without prior listener predictions, and adjust the masking to ignore these timesteps. This allows us to autoregressively predict future listener motion for arbitrary length input. \textit{No ground truth past listener motion is seen by the network at test time.}

\section{In-the-wild Conversational Dataset}
\label{sec:data}
Due to the recent COVID-19 pandemic, videotaped interviews have migrated towards teleconferencing platforms that feature a split-screen panel with the host on one side of the screen and interviewee on the other. This setup is especially advantageous for studying face-to-face communication since both individuals directly face the camera. To cover a broad range of expressions from diverse settings and people, we extract the facial motion and audio for 72 hours of videos from 6 YouTube channels. Each channel features a plethora of interviewees and hosts
from a variety of backgrounds.

We leverage a state-of-the-art facial expression extraction method, DECA~\cite{DECA:Siggraph2021}, to recover the 3D head pose and expression coefficients from in-the-wild videos. DECA estimates the pose, expression, and shape parameters according to the FLAME 3DMM~\cite{FLAME:SiggraphAsia2017}. The 3DMM defines 50 expression coefficients along with a 3D jaw rotation ($d_m = 53$), and 3D head rotation 
in Euler angles as described in Sec.~\ref{sec:motion_rep}. 
For audio, we use sound-source separation~\cite{owens2018audio} to isolate the speaker's voice.
We use these 
expressions, poses, and speaker-only audio as pseudo ground-truth 
to train our codebook (Eq.~\ref{eq:vqloss}) and prediction model (Eq.~\ref{eq:transformerloss}). 
See Appendix~\ref{appendix:data} for details. 
\textit{\textbf{We release this large-scale, novel dataset.}}

\begin{table*}\centering \footnotesize
\vspace{-.05in}
\begin{tabular}{@{}lrrrrrrcrrrrrrc@{}}\toprule 
& \multicolumn{6}{c}{Expression} & \phantom{a} & \multicolumn{6}{c}{Rotation} \\
\cmidrule{2-7} \cmidrule{9-14}
& L2 $\downarrow$ & FD $\downarrow$ & variation & SI & P-FD $\downarrow$ & PCC && L2 $\downarrow$ & FD $\downarrow$ & variation & SI & P-FD $\downarrow$ & PCC \\ 
& & ($10^3$) & & & ($10^3$) & & & & ( $10^2$) & & & ($10^2$) & \\
\midrule
\rowcolor{light-gray}
\textit{GT} & & & \textit{2.90} & \textit{2.61} & & \textit{0.09} && & & \textit{0.81} & \textit{1.96} & & \textit{0.008} \\
NN motion & 45.76 & 20.66 & 2.79 & 2.36 & 21.94 & 0.02 && 6.44 & 2.78 & 0.90 & 1.91 & 4.06 & 0.006 \\
NN audio & 52.67 & 31.98 & 2.70 & 2.41 & 33.81 & 0.02 && 7.61 & 5.54 & 0.93 & 2.01 & 6.87 & 0.007 \\
Random & 54.58 & 43.53 & 2.76 & 2.49 & 45.25 & 0.01 && 8.14 & 6.51 & 0.90 & 1.94 & 7.83 & 0.005 \\
Median & 43.18 & 64.48 & 0.00 & 0.00 & 64.77 & - && 6.35 & 15.50 & 0.00 & 0.00 & 15.50 & - \\
Mirror & 53.90 & 43.56 & 3.73 & 2.99 & 75.30 & 1.00 && 7.80 & 6.01 & 1.22 & 1.99 & 16.88 & 1.000 \\
Delayed Mirror & 53.95 & 43.79 & 3.78 & 2.88 & 76.72 & 0.98 && 7.82 & 5.99 & 1.31 & 1.93 & 16.96 & 0.999 \\
LFI \cite{jonell2020let} & 50.07 & 43.63 & 1.15 & 1.33 & 54.34 & 0.89 && 9.00 & 9.80 & 0.17 & 1.07 & 12.36 & 0.034 \\
Random Expression & 129.34 & 524.69 & 62.23 & 1.17 & 526.46 & 0.00 && 27.67 & 257.06 & 62.39 & 1.06 & 257.16 & 0.002 \\
\rowcolor{Gray}
Ours Random Walk & 52.68 & 40.45 & 1.99 & 2.26 & 42.55 & 0.01 && 7.14 & 5.74 & 0.60 & 1.37 & 7.79 & 0.001 \\
\rowcolor{Gray}
Ours & \textbf{33.16} & \textbf{3.55} & 2.01 & 2.48 & \textbf{5.15} & 0.07 && \textbf{4.75} & \textbf{0.81} & 0.62 & 1.82 & \textbf{0.87} & 0.008 \\
\bottomrule
\end{tabular}
\vspace{-.1in}
\caption{\textbf{Baselines.} Comparison against ground truth annotations (GT) on in-the-wild data. $\downarrow$ indicates lower is better; for no arrow, closer to GT is better. We bold best performances that are statistically significant. For FD and P-FD, results shown in units indicated above. 
}  
\label{tab:baselines}
\end{table*}

\section{Experiments}
We evaluate our model's ability to effectively translate the speaker's audio and motion into 
corresponding listener motion. We employ an extensive set of quantitative metrics to measure the realism, diversity, and synchrony of the listener's facial motion. Further, we perform a perceptual study to corroborate quantitative results.
All evaluations are done against the raw ground truth listener motion $y$. We discuss person-agnostic listener models in Appendix~\ref{appendix:eval}. 

\medskip \noindent \textbf{Implementation Details.}
We use $w=8$, $T=64$, $K=200$, $d_z=256$, $t=32$. We add random masking of input past listener motion. While we train on many different input speaker identities, each codebook and predictor model is trained on a specific listener (\eg~person-specific listener behavior for any speaker input). For all, we use a train/val/test split of $70\%/20\%/10\%$. Quantitative results are aggregated over all listener models. At test-time, we use nucleus sampling~\cite{holtzman2019curious}.

To improve the visual perceptibility of our results, we also train a person-specific mesh-to-pixel visualization module to directly translate 3DMM predictions to a picture of the listener (Figure~\ref{fig:teaser}). See Appendix~\ref{appendix:method}.~and video. However, since photorealistic generation is not the main focus of our work, all evaluations are done on the 3D mesh reconstructions, which are the direct outputs of our model.

\label{sec:results}


\subsection{Experimental Setup} \label{sec:baseline}

\noindent \textbf{Evaluation Metrics.}
Quantifying motion realism is a difficult problem that cannot be reduced to a single metric. We thus evaluate our predictions along multiple axes based on a composition of metrics from prior work. Our evaluation suite is based on the notion that good listeners should display (1) realistic and (2) diverse motion that is (3) synchronous with the motion of the speaker. We assess expression and rotation separately according to these three pillars: 
    
    \smallskip
    \noindent $\sbullet$ \emph{L2:} Distance to ground truth expression coefficients/pose.
    
    \smallskip
    \noindent $\sbullet$ \emph{Frechet distance for realism:} Motion realism measured by 
    distribution distance between generated and ground-truth motion sequences following~\cite{li2021learn}. We directly calculate the Frechet distance (FD)~\cite{heusel2017gans} in the expression space $\mathds{R}^{T \times d_m}$ or the head pose space $\mathds{R}^{T \times 3}$ on the full motion sequence.
    
    \smallskip
    \noindent $\sbullet$ \emph{Variation for diversity:} Variance in motion across a sequence. 
    We calculate the variance across the time series sequence of expression coefficents or 3D rotations.
    
    \smallskip
    \noindent $\sbullet$ \emph{SI for diversity:} Diverseness of predictions. As  in~\cite{zhang2020generating}, we empirically run k-means to cluster all listener expressions/rotations from training set. We compute avg.~entropy (Shannon index) of the cluster id histogram of predicted sequences. $k=15,9$ for expression, rotation, respectively.
    
    \smallskip
    \noindent $\sbullet$ \emph{Paired FD for synchrony:} Quality of listener-speaker dynamics measured by distribution distances on listener-speaker \emph{pairs} (P-FD). Calculated FD~\cite{heusel2017gans} on concatenated listener-speaker expression $\mathds{R}^{T \times (d_m+d_m)}$/ pose $\mathds{R}^{T \times (3+3)}$. 
    
    \smallskip
    \noindent $\sbullet$ \emph{PCC for synchrony:} Pearson correlation coefficient (PCC), popular metric used to quantify \emph{global} synchrony in psychology~\cite{boker_WCC, Riehle2017}. Measures how a listener covaries with a speaker over a 1D time series. We calculate
    lip curvature~\cite{ginosar2015century} to measure smile synchrony (Fig.~\ref{fig:sync}). For rotation, we measure synchrony in up/down head motion (nods).

    \smallskip
    \noindent $\sbullet$ \emph{TLCC for synchrony:} We further analyze the leader-follower relationship between our generated listeners and the input speakers by calculating the time lagged cross correlation (TLCC)~\cite{boker_WCC}. For $x \in [0,60]$ frames (up to 2s) we shift the speaker forward by $x$ frames and calculate the correlation on the delayed speaker and corresponding listener. The peak correlation indicates when the two time series are most synchronized. We also use this analysis to find the optimal delay for \textbf{Mirror Delay} baseline below.

\begin{table*}\centering \footnotesize
\vspace{-.05in}
\setlength{\tabcolsep}{3.5pt}.
\begin{tabular}{@{}lrrrrcrrrrrrcrrrrrr@{}}
& \multicolumn{4}{c}{} & \multicolumn{7}{c}{Expression} & \phantom{a} & \multicolumn{6}{c}{Rotation} \\
\midrule
& audio & motion & VQ & CA ~&& L2 $\downarrow$ & FD $\downarrow$ & variation & SI & P-FD $\downarrow$ & PCC && L2 $\downarrow$ & FD $\downarrow$ & variation & SI & P-FD $\downarrow$ & PCC\\
& \multicolumn{6}{c}{} & ($10^3$) & & & ($10^3$) & & & & ( $10^2$) & & & ($10^2$) & \\
\cmidrule{2-5} \cmidrule{7-12} \cmidrule{14-19}
\textit{GT} & \multicolumn{7}{c}{} & \textit{2.90} & \textit{2.61} & & \textit{0.09} && & & \textit{0.81} & \textit{1.96} & & \textit{0.008} \\
NoVQ $a+m$ & \cmark & \cmark & \xmark & \cmark && 36.06 & 16.60 & 0.55 & 1.69 & 18.49 & 0.05 && 4.99 & 3.64 & 0.17 & 1.21 & 3.78 & 0.006 \\
$m$ & \xmark & \cmark & \cmark & - && 38.32 & 4.10 & 1.91 & 2.46 & 5.69 & 0.12 && 5.47 & 0.96 & 0.57 & 1.80 & 1.02 & 0.009 \\
$a$ & \cmark & \xmark & \cmark & - && 39.37 & 4.11 & 1.93 & 2.47 & 5.86 & 0.06 && 5.80 & 0.91 & 0.61 & 1.78 & 0.98 & 0.007 \\
$a+m$ & \cmark & \cmark & \cmark & \xmark && 38.05 & 4.01 & 1.93 & 2.45 & 5.67 & 0.11 && 5.50 & 0.87 & 0.58 & 1.84 & 0.93 & 0.009 \\
Full & \cmark & \cmark & \cmark & \cmark && \textbf{33.16} & \textbf{3.55} & 2.01 & 2.48 & \textbf{5.15} & 0.07 && \textbf{4.75} & \textbf{0.81} & 0.62 & 1.82 & \textbf{0.87} & 0.008 \\

\bottomrule
\end{tabular}
\vspace{-.1in}
\caption{\textbf{Ablations.} Effect of ablating key components of our method. $\downarrow$ indicates lower is better; for no arrow, closer to GT is better. CA denotes cross-attention. We bold best performances that are statistically significant. For FD and P-FD, results shown in units indicated above.}
\label{tab:ablations}
\end{table*}
\medskip \noindent \textbf{Baselines.}
We compare to the following baselines:

    \noindent $\sbullet$ \textbf{NN motion:} A segment-search method commonly used for synthesis in graphics. Given an input speaker motion, we find its nearest neighbor from the training set and use its corresponding listener segment as the prediction. We found NN on the full 64-frame sequence to work better than NN on smaller subsequences that are then interpolated together.

    \smallskip
    \noindent $\sbullet$ \textbf{NN audio:} Same as above, but we find NN via audio embeddings obtained from a pretrained VGGish~\cite{hershey2017cnn} model.
    
    \smallskip
    \noindent  $\sbullet$ \textbf{Random}: Return a randomly-chosen 64-frame motion sequence of a listener from the training set.
    
    \smallskip
    \noindent $\sbullet$ \textbf{Median}: Simple yet strong baseline exploiting prior that listener is often still. Median expression/pose from train set.
    
    \smallskip
    \noindent $\sbullet$ \textbf{Mirror}: Return the speaker's motion smoothed.
    
    \smallskip
    \noindent $\sbullet$ \textbf{Delayed Mirror}: Here we mirror the speaker's smoothed motion delayed by $17$ frames ($\approx0.5$ s). While~\cite{learn2smile2017} delayed by $90$ frames, we analytically found the optimal lag according to time lagged cross correlation as discussed above.
    
    \smallskip
    \noindent $\sbullet$ \textbf{Let's Face It (LFI)~\cite{jonell2020let}}: 
    SOTA interlocutor-aware 3D avatar generation re-trained on our data. Details in Appendix~\ref{appendix:eval}.
    
    \smallskip
    \noindent $\sbullet$ \textbf{Random Expression}: 
    Walk over 3DMM space; returns a random face at each timestep.
    
    \smallskip
    \noindent $\sbullet$ \textbf{Ours Random Walk}: 
    Walk over codebook indices.

\subsection{Quantitative Results}
Table~\ref{tab:baselines} shows our proposed method outperforms all other competing methods across a variety of metrics. 
Overall, \textbf{Ours} achieves the best balance of performance across the various metrics. 
Rather than evaluating on L2 performance alone, our full suite of metrics provides a well-rounded view of the qualities of good listeners.
For instance, while \textbf{Median} performs competitively against \textbf{Ours} on L2, it suffers in terms of motion diversity (variation, SI). As a result, this baseline produces less realistic listeners, as noted by our realism metrics (FD, P-FD).
However, more variation in the facial gestures 
is not necessarily better. 
While \textbf{NN motion}, \textbf{NN audio}, and \textbf{Random} produce diversity similar to real motion, the expression synchrony (PCC) for these baselines is severely lacking. The incongruous listeners hinder the realism of the 
dyad as a whole
(P-FD). That said, a mime that mirrors the
speaker like \textbf{Mirror} and \textbf{Mirror Delay} looks uncanny due to excessive variation and synchrony. \textbf{Ours} 
delicately balances realism, diversity, and synchrony.

The weaker performance of \textbf{LFI~\cite{jonell2020let}} demonstrates the advantages of our approach. 
\textbf{LFI~\cite{jonell2020let}} was far less robust when re-trained on our in-the-wild data. Unable to learn realistic listener motion, \textbf{LFI~\cite{jonell2020let}} defaulted to mirroring the speaker, resulting in excessively high synchrony (PCC) and worse realism (FD, P-FD). Even when evaluated on the LFI~\cite{jonell2020let} dataset, ours outperforms. These results and visual comparisons in Appendix~\ref{appendix:eval}.

Additionally, we quantitatively demonstrate a major advantage of our method's VQ-VAE in learning a robust and realistic manifold of listener motion. \textbf{Ours Random Walk} is competitive against \textbf{Random}, where we sample full sequences of \emph{real motion}. It significantly outperforms  \textbf{Random Expression}, where we randomly sample static expressions and rotations at each timestep. This demonstrates that random walks along the codebook still produce realistic motion, though it may not be in sync with the speaker. 

Finally, the average TLCC calculated for \textbf{GT} and \textbf{Ours} were both $\approx 17$ frames, both reflecting an average listener response time of $\approx 0.5$s.
As mentioned above, we use this response time as the optimal delay for \textbf{Mirror Delay} baseline. See Appendix~\ref{appendix:eval} for full analysis. 

\medskip \noindent \textbf{Model Ablations.}
Table~\ref{tab:ablations} quantifies the contributions of each component of our method. In \textbf{NoVQ a+m}, we remove 
the VQ-VAE and use raw listener motion as the input and output representations. \textbf{NoVQ a+m} 
produces
unrealistic, overly smoothed sequences. Adding the VQ-VAE gives a significant performance boost, which further confirms the importance of the codebook in generating realistic motion. Furthermore, we demonstrate that utilizing both audio and motion as input \textbf{a+m} via concatenation slightly improves performance over using just one or the other (\textbf{a} and \textbf{m}). However, \textbf{Ours} achieves a more substantial improvement when combining both modalities via cross-attention (CA). See Appendix~\ref{appendix:eval} for details of ablation architectures.

\subsection{Qualitative Results}
To corroborate our quantitative results and gain insight into how our synthesized listeners perceptually compare to real motion
, we conducted an A/B test on Amazon Mechanical Turk. Since all quantitative trends were consistent across all listener identities, we randomly chose a single identity for the evaluation. 
We visualized listener motion using videos of grayscale 3D facial meshes. 

Participants watched a series of video pairs. In each pair, one video was generated from our model; the other was produced by an ablation 
or a baseline.
Participants were then asked to identify the video containing the listener that looks like they are listening and paying \emph{more} attention to the speaker.
Videos of 8 seconds each of resolution $849 \times 450$ (downsampled from $1132 \times 600$ in order to fit two videos vertically stacked on different screen sizes) were shown, and after each pair, participants were given unlimited time to respond. 
Since the most tell-tale moments for when a listener is truly listening are during defining moments (speaker tells a joke, shares a sad story, \etc.) that illicit strong responses, we manually curated such notable moment sequences from our held-out test data. We then randomly sampled $50$ from these sequences and predicted a corresponding listener 3D facial motion sequence using each method. For every test sequence, each A/B comparison was made by 3 evaluators.

We compared our strongest baseline \textbf{NN motion} and ablation \textbf{a+m} to our proposed model and recorded the percentage of times our method was preferred over the baseline models or vice versa. \textbf{Ours} significantly outperformed. $75.3\%$ of the total 150 evaluators preferred \textbf{Ours} over \textbf{NN}, and $71.1\%$ preferred \textbf{Ours} over \textbf{a+m}. These statistics reflect the quantitative trends in Table~\ref{tab:ablations}.~Furthermore, in a comparison against avatars rendered from ground truth listeners, evaluators preferred \textbf{Ours} $50.1\%$ of the time. This \textit{highlights the perceptual realism of our predicted listener motion}.

\section{Discussion}
\label{sec:discussion}
In this work, we explored the synchronicity of motion between a speaker and a listener. To this end, we employ a motion-audio cross-attention transformer to handle the multiple modalities of speaker inputs. Furthermore, we enable non-deterministic motion synthesis with a VQ-VAE. Trained on a novel, in-the-wild dataset of dyadic conversations, our method autoregressively outputs convincing 3D listener facial motion that correlate with a given speaker. 

While videotaped teleconferencing data lends itself to data collection, it has inherent limitations (\eg no eye contact, time delays introduced by remote connections, \etc). A future direction would be to apply this study to in-person conversations, which would allow us to incorporate gaze. Furthermore, as we only model listener motion in response to a speaker, modeling the full dyadic cycle of back-and-forth effect remains for future work.
While our goal is to understand conversational dynamics, we discuss concerns for misuse of this technology in Appendix.~Please see Appendix per-listener results, implementation details, ablation architectures, multiple mode output evaluation,~\etc.

\medskip
\noindent \textbf{Acknowledgements.} The authors would like to thank Justine Cassell, Alyosha Efros, Alison Gopnik, Jitendra Malik, and the Facebook FRL team for many insightful conversations and comments. Dave Epstein and Karttikeya Mangalam for Transformer advice. Ruilong Li and Ethan Weber for technical support. The work of Ng and Darrell is supported in part by DoD including DARPA's XAI, LwLL, Machine Common Sense and/or SemaFor programs, which also supports Hu and Li in part, as well as BAIR's industrial alliance programs.
Ginosar's work is funded by the NSF under Grant \# 2030859 to the Computing Research Association for the CIFellows Project. Parent authors would like to thank their children for the daily reminder that they should learn how to listen.

{\small
\bibliographystyle{ieee_fullname}
\bibliography{main}
}

\clearpage

{\Large{\textbf{Appendix}}}
\appendix

\section{Related Works} \label{appendix:rw}
\noindent \textbf{Interactional Motion in Human Communication.}
Humans are able to enter into synchronous negotiations with others from early infancy~\cite{infant_itersubjectivity_1979}. This interaction primitive is so central to human communication that infants believe anything that behaves synchronously with them is an independent agent even if it not human-like in appearance~\cite{whose_gase_1998}. The infant-mother dyad of affective synchrony of motion~\cite{Messinger2009,movellan2015} is of particular importance.
Face-to-face infant-mother synchrony has long-term consequences---it can predict a child's temperament years later~\cite{feldman1999mother}.

The study of face-to-face communication has been classically hindered not only due to lack of data, but also due to computational methods that could analyze it.
The earliest studies of motion patterns in dyadic conversations involved manual analysis of videotaped data~\cite{condon_ogston_1966,kendon1970movement}.
Condon and Ogston~\cite{condon_ogston_1966} describe an interactional synchrony
where the motion of the listener flows in rhythm with the speech and motion of the speaker. Kendon~\cite{kendon1970movement} extends their study of head motion to that of the upper body, manually transcribing motion patterns of speakers and listeners in video recordings.
While several coding systems for interaction synchrony~\cite{izard1979maximally,tronick1980monadic,cohn1987mother}
were later developed, they all still required laborious manual annotation efforts. 
With the advent of modern computer vision, frame-differencing methods provided much-needed automation~\cite{Paxton2013}. 

Several works computationally demonstrate the existence of distinctive interactional motion in face-to-face interactions. 
Riehle \etal 
employ correlation analysis~\cite{boker_WCC} on electromyography (EMG) signals of recorded muscle activations and suggest that people typically synchronize their smiles with those of their interlocutors within $1$ second. Other methods for detecting synchrony use facial 2D~\cite{chu_unsupervised_2015} or full-body 3D~\cite{gaziv2017reduced} keypoint detectors.
Our method leverages the existence of interactional syncrhony to learn realistic facial motion of a listener in response to a speaker's speech and motion by training on a large dataset of in-the-wild conversations.

\section{Method} \label{appendix:method}

\noindent \textbf{VQ-VAE Details.}
The VQ-VAE is composed of 3 convolutional layers of kernel size 5, stride 1, padding 2. Each convolutional layer is followed by a max pooling operation. Given a sequence length 32, following the convolutional layers, we have a sequence length 4 (window size $w=8$. We pass this bottlenecked sequence through a Transformer of hidden size 512, number of heads 8, number of layers 12. The positional encoding is learned. We train the VQ-VAE on sequences of length 32 for 1000 epochs ($\approx 1$ day on 8 GPU's) with a learning rate of 2.0 with 4,000 warm-up steps. We optimize using Adam with a batch size of 32. Train/val/test split is 70/20/10. We then use the frozen model downstream to quantize the listener inputs to the Predictor. During test time, the average L2 error induced by quantization on the test set is 11.32 on expression and 1.02 on rotation. 

\begin{figure}[t]
\begin{center}
\includegraphics[width=0.9\linewidth]{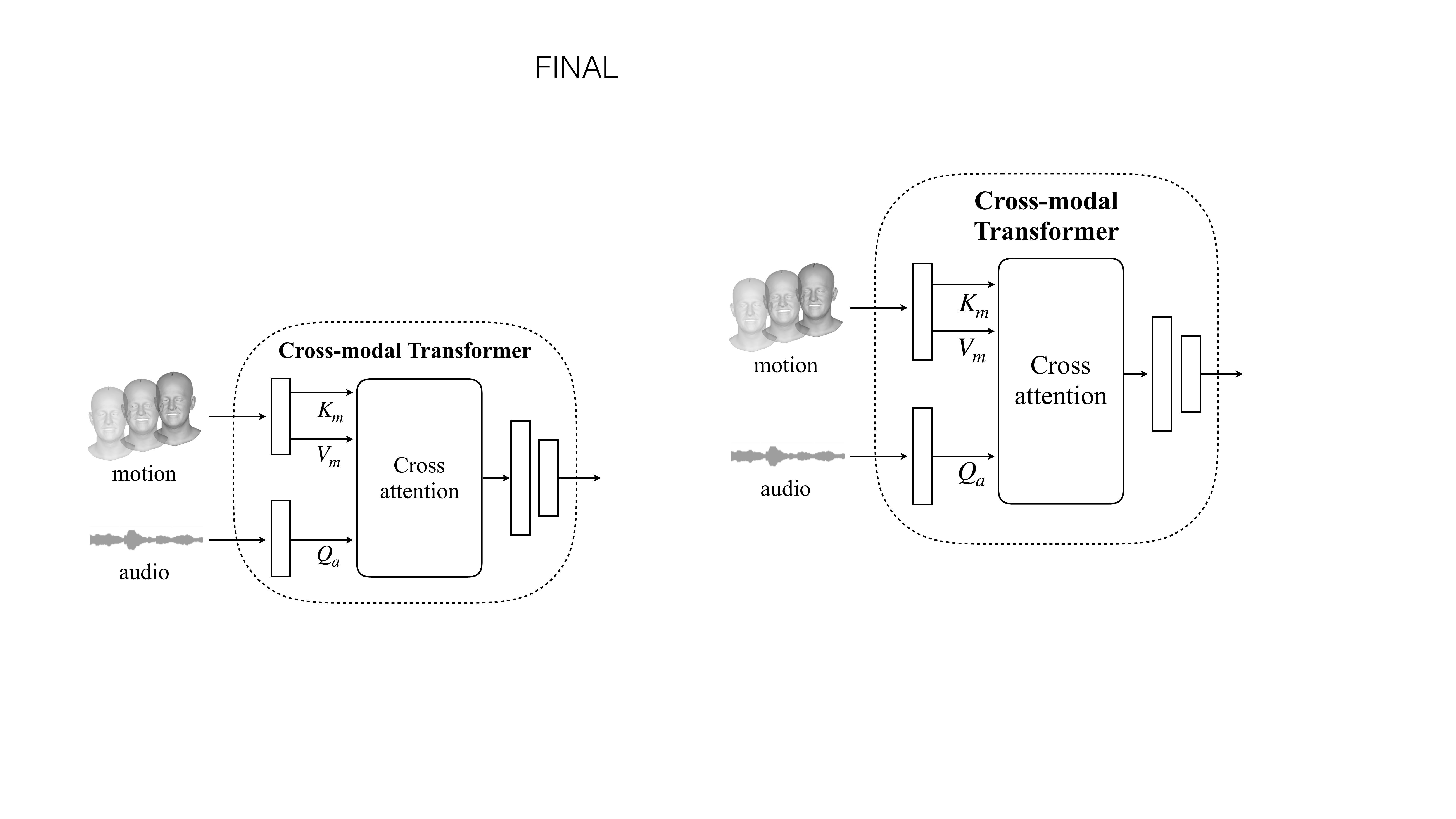}
\end{center}
    \vspace{-1.5em}
  \caption{\textbf{Cross-modal transformer detail.} We combine the different modalities of the \textit{speaker's} input via a cross modal transformer. Each modality first passes through a linear layer. Then, a stack of cross attention blocks treats the audio as queries and the speaker's motion as the keys and values. Finally, 1D convolutions temporally downsample the output, to match the temporal extent of the quantized listener input.}
\label{fig:cross-modal-small-framed}
\end{figure}

\medskip
\noindent \textbf{Cross-modal Transformer Details.}
The cross-modal transformer takes as input the raw motion representation mentioned in the main text (Sec.~3.1). To process the audio, we use the audio processing library, librosa, to process the audio at a sampling rate of 16000, and to obtain the melspectrogram. This results in a sequence 4 times longer than the motion sequence extracted at 30fps. As a result, we apply max pooling to downsample the audio to a sequence length that matches the motion sequence. We feed the downsampled audio and the motion independently through a Linear layer for each modality to obtain their respective projected embeddings. These are then fed into the cross-modal transformer shown in Figure.~\ref{fig:cross-modal-small-framed}.
The cross-modal transformer is composed of a transformer of hidden size 1024, number of heads 8, number of layers 12. Following the transformer are 3 convolutional layers of kernel size 5, stride 1, padding 2. Each convolutional layer is followed by a max pooling operation that temporally downsamples the speaker embedding of length 32 to match the size of the listener embedding ($\tau = 4$). The positional encoding of the transformer is learned. 

While we first attempted to not downsampling the speaker with convolutional layers, feeding the full 32 length sequence into the Predictor as the speaker input resulted in empirically worse results across the board. We suspect this is due to overly long conditioning and a temporal mismatch of the speaker's much longer token sequence with the listener's shorter quantized sequence. This would require the network to learn the temporal alignment of the two different length sequences on it's own. 

\medskip
\noindent \textbf{Predictor Details.}
Our predictor is composed of a transformer with hidden size 200, number of heads 10, and number of layers 5. To convert the sequence of listener indices $\in \mathds{R}^{\tau \times 1}$ to a embedding $\in \mathds{R}^{\tau \times d_k}$ that matches the size of the speaker embedding, we embed the indices using the Embedding function in Pytorch. We then concatenate the output of the cross modal speaker embedding with the listener embedding to get a sequence $\in \mathds{R}^{2\tau \times d_k}$, which serves as input to the predictor.

Since the Transformer is a set-to-set operation, it outputs a sequence of length $2\tau$ matching the input. During training we take the first 4 indices of the output and discard the remainder. The additional 3 indices are used as a temporal regularizer similar to [41]. We train the predictor jointly with the cross-modal transformer for 1000 epochs ($\approx 12$ hours on 8 GPU's) with a learning rate of 0.01 with 4,000 warm-up steps. During test-time, we take only the first index.

To facilitate autoregressive learning, we mask out the attention on portions of the past listener input. Given a listener sequence of length $\tau$, we end up with an attention matrix $\tau \times \tau$. During training time, we generate a random number $x = [0,\tau]$ $50\%$ of the time and set the attention for rows above $\tau$, representing time-steps preceding $x$, negligibly low. For the remaining $50\%$, we do not mask anything out. During test time, we start by masking everything out and then gradually reduce the mask as we autoregressively predict the output. This ensures we only see past listeners we predict.

\medskip
\noindent \textbf{Visualization.}
To improve the visual perceptibility of our results, we also train a person-specific 3DMM-to-video translation network.
We adopt the state-of-the-art video-to-video synthesis method \cite{wang2018vid2vid} to translate the grey scale 3DMM rendering results into full frames of a photo-realistic target video, in which the target listener mimics the facial expression and the head motion of the grey scale visualization. Our network learns to simulate the static background and the entire listener, where the face region is conditioned on the 3DMM rendering, while other components, such as hair and torso are compiled with the head pose. During the training session, the training data is extracted from a single video clip. The ground truth targets are the frames of the listener in the video, and the sources are the renderings of the corresponding 3DMM predictions.  



\section{In-the-wild Conversational Dataset} \label{appendix:data}
Pursuant to guidance from our local IRB we have determined that there is no non public PII and no human subjects under 45 CFR 46 in our dataset.

\medskip
\noindent \textbf{Dataset Preprocessing.}
All frames are extracted at 30 fps. 
We pre-process the video data by automatically removing irrelevant segments and annotating listener versus speaker splits.
While these interview videos often contain views of both the host and the guest in a split-screen format, the view often switches back to a single individual, or to an inserted picture. We employed an off-the-shelf facial detection method (DECA~[Feng et al.~2021]) to automatically find all relevant segments of two conversing individuals -- e.g.~removing parts of the video with less than two faces detected, with a face detected in the center of the screen and an extraneous false positive in the background, or with a face that was static for an extended period. Furthermore, to minimize the noisiness of the extracted pseudo-ground truth annotations, we removed frames where most of the detected 2D keypoints estimated by DECA were missing, and sequences with a sudden extreme movement.

To ensure that we feed speaker segments as input and use listener segments as pseudo ground truth labels, we additionally extract listener versus speaker splits for each video. We automatically detect the splits by using active speaker identification~[Chung et al.~2016] in conjunction with sound source separation~[Owens et al.~2018]. We found that using either method independently led to noisy speaker predictions for different reasons. Combining the two in a voting-based approach led to more reliable splits. We removed sequences where both individuals speak at the same time, but kept sequences where the listener utters a few words (e.g.~to show agreeance).

\section{Evaluation} \label{appendix:eval}

\noindent \textbf{Ablation Architectures.}
\begin{figure}[t]
\begin{center}
\includegraphics[width=0.9\linewidth]{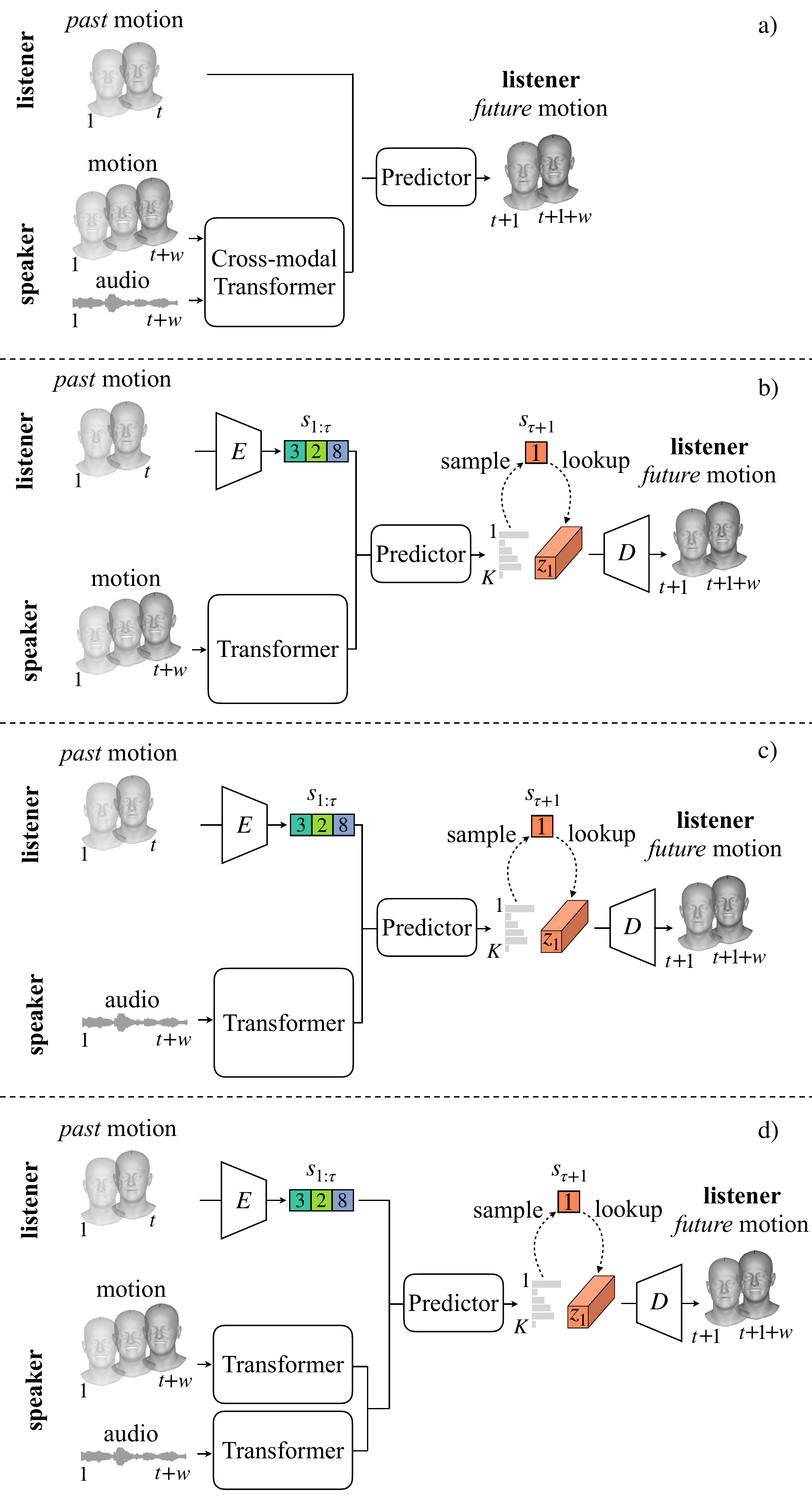}
\end{center}
    \vspace{-1.5em}
  \caption{\textbf{Ablation details.} The ablations performed in Table 2. a) \textbf{NoVQ a+m}: We remove the VQ-VAE component, taking as input the raw listener and outputting the raw listener. We take as input only the b) motion \textbf{m} or the c) audio \textbf{a} from the speaker, replacing the cross-modal transformer with a normal transformer for the single-modal case. d) \textbf{a+m}: Rather than using a cross-modal transformer, we simply pass each modality through a transformer and perform fusion via concatenation. }
\label{fig:ablations}
\end{figure}
In Table 1 of the main paper, we perform multiple ablations to test the validity of our method's components. We further describe and sketch out the ablation architectures as shown in Figure~\ref{fig:ablations}.

\medskip
\noindent \textbf{LFI~[Jonell et al.~2020] Implementation Details.} 
Since we could not get access to their trained model nor their video dataset, we needed to retrain their method on our video dataset. Training on our dataset would also allow us to compare both methods more fairly, since there is a large distribution shift between their dataset and ours. Rather than changing their model, we use the same setup provided (inputs, configs, etc.~) and simply retrain on our dataset. We use the config including both audio and motion for the listener and speaker, which may provide their model with slightly more context as described in Section 2 of the main paper. We use the same optimizer, learning rates, and training schedule as defined by the authors of the paper. We train the model for 3 days on 8 GPU's.

\medskip
\noindent \textbf{Individual Styles of Listening Motion.}
In the main paper, we show the quantitative results averaged over 4 listeners (Persons A, B, C, and D).
Person specific results are shown in Tables~\ref{tab:conan},~\ref{tab:fallon},~\ref{tab:trevor}, and~\ref{tab:stephen} respectively. The trends in each of the different person-specific cases parallel those of the averaged case. \textbf{Ours} outperforms all baselines and ablations in all.
Please note, we excluded two of our listeners due to the fact that they include guest hosts who serve the role in many videos. Therefore we did not have enough data to learn a person-specific model with the remaining data. However, we will still release these portions of the data. 

The decision to train different models per-person comes from the idea that each person has a characteristic way of listening. By training person-specific models, we can capture these more fine-grain characteristic details of listener motion, which is difficult to do when considering all identities at once. 


\begin{table*}\centering \footnotesize
\vspace{-.05in}
\begin{tabular}{@{}lrrrrrrcrrrrrrc@{}}\toprule 
& \multicolumn{6}{c}{Expression} & \phantom{a} & \multicolumn{6}{c}{Rotation} \\
\cmidrule{2-7} \cmidrule{9-14}
& L2 $\downarrow$ & FD $\downarrow$ & variation & SI & P-FD $\downarrow$ & PCC && L2 $\downarrow$ & FD $\downarrow$ & variation & SI & P-FD $\downarrow$ & PCC \\ 
& & ($10^3$) & & & ($10^3$) & & & & ( $10^2$) & & & ($10^2$) & \\
\midrule
\textit{GT} & & & \textit{3.13} & \textit{2.62} & & \textit{0.02} && & & \textit{0.98} & \textit{2.05} & & \textit{0.0104} \\
NN motion & 45.51 & 6.54  & 2.86  & 2.86  & 8.45  & 0.01  && 8.22  & 5.96  & 1.09  & 2.02  & 11.00 & 0.0083 \\
NN audio & 45.70 & 7.41  & 2.65  & 2.56  & 9.38  & 0.00  & &7.11  & 6.47  & 0.95  & 2.04  & 11.58 & 0.0099 \\
Random & 41.43 & 7.55  & 2.81  & 2.52  & 8.87  & 0.01  && 9.36  & 8.78  & 0.92  & 1.98  & 13.86 & 0.0011 \\
Median & 28.79 & 29.30 & 0.00  & 0.00  & 29.30 & -     && 8.04  & 23.52 & 0.00  & 0.00  & 23.52 & -      \\
Mirror & 42.97 & 7.35  & 3.34  & 2.99  & 41.32 & 1.00  && 8.26  & 5.42  & 1.31  & 2.10  & 19.00 & 1.0000 \\
Delayed Mirror & 42.69 & 7.33  & 3.46  & 2.88  & 40.23 & 0.97  && 8.32  & 5.75  & 1.43  & 1.86  & 19.23 & 0.9924 \\
LFI~[34] & 41.89 & 8.68  & 1.09  & 1.33 & 39.86 & 0.65  && 9.10  & 7.96  & 0.11  & 1.28  & 18.02 & 0.0139 \\
Random Expression & 126.89& 496.27& 69.65 & 1.17  & 498.22& 0.00  && 28.63 & 270.62& 69.83 & 0.90  & 270.72& 0.0005 \\
No VQ $a+m$ & 36.76 & 7.23  & 1.11  & 2.28  & 8.60  & 0.04  && 4.33  & 3.34  & 0.24  & 0.96  & 3.37  & 0.0069 \\
$m$ & 38.04 & 3.59  & 1.95  & 2.48  & 4.89  & 0.04  && 5.91  & 0.79  & 0.69  & 1.87  & 0.85  & 0.0106 \\
$a$ & 39.21 & 3.55  & 2.03  & 2.58  & 5.21  & 0.00  && 6.32  & 0.60  & 0.81  & 1.70  & 0.68  & 0.0085 \\
$a+m$ & 37.92 & 3.33  & 2.00  & 2.47  & 4.99  & 0.03  && 6.02  & 0.60  & 0.79  & 1.92  & 0.65  & 0.0108 \\
\rowcolor{Gray}
Ours Random Walk & 55.26 & 7.36  & 1.76  & 2.40  & 41.42 & 0.01  && 7.36  & 5.49  & 0.63  & 1.35  & 13.54 & 0.0013 \\
\rowcolor{Gray}
Ours & 34.94 & 3.20  & 2.08  & 2.50  & 4.73  & 0.02  && 5.21  & 0.54  & 0.72  & 1.88  & 0.62  & 0.0093 \\
\bottomrule
\end{tabular}
\vspace{-.1in}
\caption{\textbf{Person A.} Comparison against ground truth annotations (GT) on in-the-wild data. $\downarrow$ indicates lower is better; for no arrow, closer to GT is better. We bold best performances that are statistically significant. For FD and P-FD, results shown in units indicated above. 
}  
\label{tab:conan}
\end{table*}

\begin{table*}\centering \footnotesize
\vspace{-.05in}
\begin{tabular}{@{}lrrrrrrcrrrrrrc@{}}\toprule 
& \multicolumn{6}{c}{Expression} & \phantom{a} & \multicolumn{6}{c}{Rotation} \\
\cmidrule{2-7} \cmidrule{9-14}
& L2 $\downarrow$ & FD $\downarrow$ & variation & SI & P-FD $\downarrow$ & PCC && L2 $\downarrow$ & FD $\downarrow$ & variation & SI & P-FD $\downarrow$ & PCC \\ 
& & ($10^3$) & & & ($10^3$) & & & & ( $10^2$) & & & ($10^2$) & \\
\midrule
\textit{GT} & & & \textit{3.24} & \textit{2.62} & & \textit{0.06} && & & \textit{0.95} & \textit{1.96} & & \textit{0.0091} \\
NN motion & 55.71   & 22.62  & 3.13  & 2.34 & 23.78  & 0.01 &  & 7.46  & 1.70   & 0.95  & 2.02 & 1.72   & 0.0080 \\
NN audio & 58.74   & 30.70  & 2.83  & 2.43 & 32.51  & 0.03 &  & 8.47  & 3.28   & 0.96  & 2.03 & 3.35   & 0.0087 \\
Random & 62.06   & 48.12  & 2.69  & 2.46 & 49.76  & 0.00 &  & 8.11  & 3.41   & 0.88  & 1.97 & 3.45   & 0.0026 \\
Median & 51.37   & 88.27  & 0.00  & 0.00 & 88.27  & -    &  & 6.21  & 15.27  & 0.00  & 0.00 & 15.27  & -      \\
Mirror & 56.24   & 42.36  & 4.04  & 3.08 & 93.48  & 1.00 &  & 7.99  & 7.29   & 0.89  & 2.06 & 15.72  & 1.0000 \\
Delayed Mirror & 56.30   & 42.71  & 4.16  & 2.88 & 75.24  & 0.98 &  & 8.00  & 7.04   & 1.03  & 2.05 & 15.83  & 0.9984 \\
LFI~[34] & 56.44   & 51.37  & 1.10  & 1.55 & 51.86  & 0.95 &  & 9.94  & 14.99  & 0.21  & 1.00 & 15.08  & 0.0316 \\
Random Expression & 153.35  & 726.68 & 85.27 & 1.28 & 728.64 & 0.00 &  & 32.58 & 351.74 & 85.47 & 0.94 & 351.87 & 0.0026 \\
No VQ $a+m$ & 35.65   & 26.30  & 0.58  & 1.67 & 28.87  & 0.02 &  & 5.65  & 3.66   & 0.27  & 1.58 & 4.02   & 0.0069 \\
$m$ & 38.05   & 4.57   & 1.58  & 2.35 & 6.38   & 0.09 &  & 5.60  & 1.04   & 0.59  & 1.86 & 1.14   & 0.0095 \\
$a$ & 39.13   & 4.84   & 1.63  & 2.33 & 6.61   & 0.04 &  & 5.78  & 1.30   & 0.66  & 1.91 & 1.39   & 0.0083 \\
$a+m$ & 38.28   & 4.46   & 1.59  & 2.35 & 6.39   & 0.07 &  & 5.64  & 1.16   & 0.60  & 1.90 & 1.24   & 0.0105 \\
\rowcolor{Gray}
Ours Random Walk & 50.65 & 44.39  & 1.70  & 2.07 & 41.53  & 0.01 &  & 7.37  & 3.39   & 0.63  & 1.43 & 3.44   & 0.0015 \\
\rowcolor{Gray}
Ours & 32.25   & 4.24   & 1.91  & 2.37 & 5.79   & 0.04 &  & 4.34  & 0.95   & 0.76  & 1.90 & 1.06   & 0.0089 \\
\bottomrule
\end{tabular}
\vspace{-.1in}
\caption{\textbf{Person B.} Comparison against ground truth annotations (GT) on in-the-wild data. $\downarrow$ indicates lower is better; for no arrow, closer to GT is better. We bold best performances that are statistically significant. For FD and P-FD, results shown in units indicated above. 
}  
\label{tab:fallon}
\end{table*}

\begin{table*}\centering \footnotesize
\vspace{-.05in}
\begin{tabular}{@{}lrrrrrrcrrrrrrc@{}}\toprule 
& \multicolumn{6}{c}{Expression} & \phantom{a} & \multicolumn{6}{c}{Rotation} \\
\cmidrule{2-7} \cmidrule{9-14}
& L2 $\downarrow$ & FD $\downarrow$ & variation & SI & P-FD $\downarrow$ & PCC && L2 $\downarrow$ & FD $\downarrow$ & variation & SI & P-FD $\downarrow$ & PCC \\ 
& & ($10^3$) & & & ($10^3$) & & & & ( $10^2$) & & & ($10^2$) & \\
\midrule
\textit{GT} & & & \textit{2.84} & \textit{2.60} & & \textit{0.21} && & & \textit{0.95} & \textit{2.00} & & \textit{0.0067} \\
NN motion & 43.26   & 27.07  & 2.84  & 2.21 & 28.26  & 0.06 &  & 6.72  & 1.56   & 1.17  & 1.96 & 1.62   & 0.0056 \\
NN audio & 52.30   & 38.20  & 2.71  & 2.33 & 39.85  & 0.03 &  & 7.87  & 2.70   & 0.97  & 2.07 & 2.75   & 0.0059 \\
Random & 56.25   & 49.74  & 2.74  & 2.56 & 51.62  & 0.01 &  & 7.70  & 2.81   & 0.79  & 2.00 & 2.89   & 0.0151 \\
Median & 44.24   & 64.41  & 0.00  & 0.00 & 64.41  & -    &  & 5.81  & 13.34  & 0.00  & 0.00 & 13.34  & -      \\
Mirror & 57.22   & 56.22  & 3.56  & 2.88 & 90.40  & 1.00 &  & 8.10  & 3.91   & 1.41  & 2.07 & 16.90  & 1.0000 \\
Delayed Mirror & 57.41   & 56.39  & 3.57  & 2.89 & 89.22  & 0.98 &  & 8.13  & 3.84   & 1.51  & 2.06 & 16.95  & 0.9678 \\
LFI~[34] & 51.50   & 50.35  & 1.20  & 1.34 & 60.89  & 0.98 &  & 9.01  & 7.72   & 0.15  & 1.01 & 7.76   & 0.0752 \\
Random Expression & 112.84  & 393.95 & 42.69 & 1.03 & 395.52 & 0.00 &  & 23.62 & 182.36 & 42.80 & 1.25 & 182.44 & 0.0017 \\
No VQ $a+m$ & 37.25   & 22.92  & 0.24  & 0.45 & 24.51  & 0.12 &  & 5.34  & 5.98   & 0.05  & 0.70 & 6.03   & 0.0062 \\
$m$ & 38.98   & 5.05   & 2.10  & 2.40 & 6.39   & 0.22 &  & 5.38  & 0.93   & 0.62  & 1.78 & 0.96   & 0.0068 \\
$a$ & 40.89   & 5.05   & 2.04  & 2.40 & 6.74   & 0.11 &  & 5.57  & 0.81   & 0.55  & 1.79 & 0.86   & 0.0055 \\
$a+m$ & 38.07 & 4.96   & 2.05  & 2.42 & 6.48   & 0.25 &  & 5.32  & 0.81   & 0.55  & 1.79 & 0.85   & 0.0075 \\
\rowcolor{Gray}
Ours Random Walk & 53.46   & 44.92  & 2.43  & 2.25 & 46.89  & 0.00 &  & 7.01  & 3.39   & 0.52  & 1.18 & 3.42   & 0.0006 \\
\rowcolor{Gray}
Ours & 34.41   & 4.02   & 1.96  & 2.46 & 5.56   & 0.18 &  & 4.80  & 0.64   & 0.63  & 1.82 & 0.67   & 0.0060 \\
\bottomrule
\end{tabular}
\vspace{-.1in}
\caption{\textbf{Person C.} Comparison against ground truth annotations (GT) on in-the-wild data. $\downarrow$ indicates lower is better; for no arrow, closer to GT is better. We bold best performances that are statistically significant. For FD and P-FD, results shown in units indicated above. 
}  
\label{tab:stephen}
\end{table*}

\begin{table*}\centering \footnotesize
\vspace{-.05in}
\begin{tabular}{@{}lrrrrrrcrrrrrrc@{}}\toprule 
& \multicolumn{6}{c}{Expression} & \phantom{a} & \multicolumn{6}{c}{Rotation} \\
\cmidrule{2-7} \cmidrule{9-14}
& L2 $\downarrow$ & FD $\downarrow$ & variation & SI & P-FD $\downarrow$ & PCC && L2 $\downarrow$ & FD $\downarrow$ & variation & SI & P-FD $\downarrow$ & PCC \\ 
& & ($10^3$) & & & ($10^3$) & & & & ( $10^2$) & & & ($10^2$) & \\
\midrule
\textit{GT} & & & \textit{2.39} & \textit{2.59} & & \textit{0.06} && & & \textit{0.36} & \textit{1.83} & & \textit{0.0063} \\
NN motion & 38.55  & 26.41  & 2.34  & 2.05 & 27.28  & 0.01 &  & 3.36  & 1.89   & 0.38  & 1.63 & 1.91   & 0.0021 \\
NN audio & 53.92  & 51.60  & 2.59  & 2.38 & 53.48  & 0.02 &  & 6.99  & 9.75   & 0.84  & 1.88 & 9.83   & 0.0025 \\
Random & 58.59  & 68.71  & 2.81  & 2.41 & 70.77  & 0.00 &  & 7.40  & 11.04  & 1.01  & 1.83 & 11.12  & 0.0007 \\
Median & 48.31  & 75.92  & 0.00  & 0.00 & 75.92  & -    &  & 5.34  & 9.85   & 0.00  & 1.03 & 9.85   & -      \\
Mirror & 59.19  & 68.31  & 3.99  & 3.00 & 76.00  & 1.00 &  & 6.85  & 7.41   & 1.25  & 1.76 & 15.92  & 1.0000 \\
Delayed Mirror & 59.40  & 68.71  & 3.94  & 2.89 & 102.19 & 1.00 &  & 6.83  & 7.34   & 1.25  & 1.76 & 15.83  & 0.9984 \\
LFI~[34] & 50.43  & 64.13  & 1.21  & 1.09 & 64.75  & 0.96 &  & 7.94  & 8.53   & 0.19  & 1.00 & 8.59   & 0.0169 \\
Random Expression & 124.29 & 481.86 & 51.31 & 1.19 & 483.46 & 0.00 &  & 25.85 & 223.54 & 51.45 & 1.16 & 223.61 & 0.0020 \\
No VQ $a+m$ & 34.57  & 9.94   & 0.27  & 2.04 & 11.99  & 0.04 &  & 4.67  & 1.60   & 0.12  & 1.59 & 1.69   & 0.0033 \\
$m$ & 38.22  & 3.18   & 2.03  & 2.62 & 5.05   & 0.10 &  & 4.98  & 1.06   & 0.35  & 1.70 & 1.14   & 0.0065\\
$a$ & 38.25  & 3.00   & 2.02  & 2.58 & 4.86   & 0.03 &  & 5.53  & 0.92   & 0.39  & 1.70 & 0.99   & 0.0066 \\
$a+m$ & 37.99  & 3.24   & 2.07  & 2.58 & 4.86   & 0.08 &  & 5.03  & 0.89   & 0.40  & 1.75 & 0.97   & 0.0066 \\
\rowcolor{Gray}
Ours Random Walk & 51.36  & 65.16  & 2.06  & 2.34 & 40.39  & 0.01 &  & 6.83  & 10.70  & 0.63  & 1.51 & 10.78  & 0.0010 \\
\rowcolor{Gray}
Ours & 31.03  & 2.74   & 2.07  & 2.60 & 4.51   & 0.04 &  & 4.69  & 1.07   & 0.35  & 1.70 & 1.13   & 0.0059 \\
\bottomrule
\end{tabular}
\vspace{-.1in}
\caption{\textbf{Person D.} Comparison against ground truth annotations (GT) on in-the-wild data. $\downarrow$ indicates lower is better; for no arrow, closer to GT is better. We bold best performances that are statistically significant. For FD and P-FD, results shown in units indicated above. 
}  
\label{tab:trevor}
\end{table*}

\medskip
\noindent \textbf{Listener-agnostic modeling.}
While in the paper we focus on listener-specific models, experiments show that our method can be extended to new listeners. In a listener-agnostic setup, we train across many listeners and test on held-out listeners and speakers. All baselines in Tab.~1 are performed on listener-specific data, but when recomputed in an agnostic setup, the relative ordering of baselines does not change, with \textbf{Ours-agnostic} first (FD: $30.01$, PFD: $31.36$) and \textbf{NN motion-agnostic} as second best (FD: $57.29$, PFD: $57.91$)(lower is better).
\emph{\textbf{Even in a listener agnostic setup, our approach outperforms existing baselines.}}
While our person-agnostic model does well, person-specific modeling better captures individualistic listening styles and mannerisms~[20].

\medskip
\noindent \textbf{Evaluation on LFI [34] dataset.}
As [34] only provides facial annotations (no audio/video), we did not include an evaluation against this data in the main paper. Still, we can use their data in a motion-only setting with no speaker audio input. As the models for LFI are not publicly available, We retrained \textbf{LFI} and our ablated model \textbf{m} using their facial annotations and train/test splits. On their data, our \textbf{m} (FD: 1.88, PFD: 2.12, var: 1.82) outperforms \textbf{LFI} (FD: 2.97, PFD: 3.10, var: 1.01).

\medskip
\noindent \textbf{Metrics Details.}
Here, we further define our metrics. We denote $x \in \mathds{R}^{N \times T \times F}$ as the input speaker motion sequence, $y \in \mathds{R}^{N \times T \times F}$ as the ground truth listener motion sequence and $\hat{y} \in \mathds{R}^{N \times T \times F}$ as our prediction. $N$ is the number of test sequences, $T$ length of the sequence, and $F$ feature dimension.
\begin{itemize}
    \item \textbf{L2}: $ d = \norm{y - \hat{y}}$
    
    \item \textbf{FD}: $d^2 = {|\mu_{\hat{y}} - \mu_{y}|}^2 + tr(\Sigma_{\hat{y}} + \Sigma_{y} - 2 (\Sigma_{\hat{y}} \Sigma_{y})^{1/2})$, where $\Sigma$ is the covariance matrix.
    
    \item \textbf{variation}:  We calculate the variance along the axis representing the temporal extent $T$ of the sequence. We then average over $B$ and $F$.
    
    \item \textbf{SI}: We empirically perform kmeans clustering to cluster the expression and rotation separately as defined in the main paper. We then compute the entropy (Shannon index) of the cluster ID histogram of all the samples. We report the entropy in the Tables. 
    
    \item \textbf{P-FD}: We use the exact same equation as above \textbf{FD}, but rather than using $y$ and $\hat{y}$, we replace them with the concatenation $y \bigoplus x$ and $\hat{y} \bigoplus x$ respectively.
    
    \item \textbf{PCC}: The correlation coefficent is given as: 
    \begin{equation}
        r = \frac{\Sigma(x_t - \bar{x}) (y_t - \bar{y})}{\sqrt{\Sigma (x_t - \bar{x})^2 \Sigma(y_i - \bar{y})^2}},
    \end{equation}
    where $t \in [0,T]$ is the timestep, and $\bar{x}$ denotes the mean of  $x$. 
\end{itemize}

\medskip
\noindent \textbf{Multiple Modes of Output Analysis.}
\begin{figure}[t]
\begin{center}
\includegraphics[width=\linewidth]{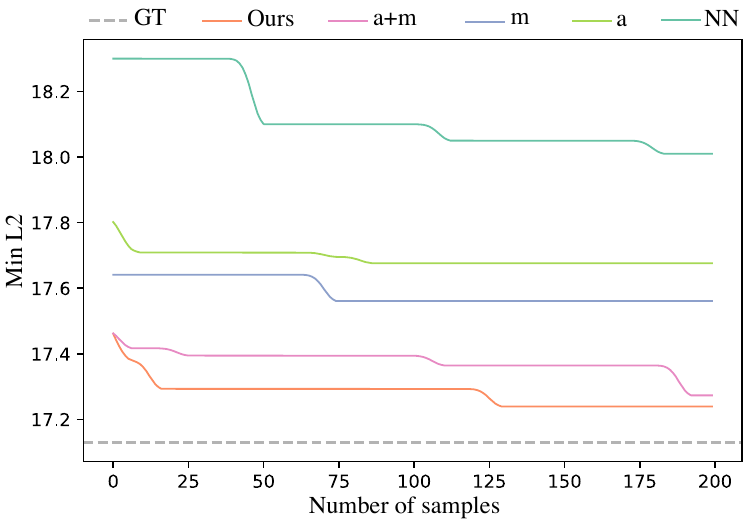}
\end{center}
    \vspace{-0.8em}
   \caption{For each method, we feed in the same 256 length test sequences, and sample $x \in [1,200]$ output listener motion sequences. From the $x$ sampled trajectories, we calculate the avg.~minimum L2 distance to ground truth. For \textbf{NN}, we sample the top 200 64-length sequences. The loss from quantizing \textbf{GT} is shown ($\text{L2}=17.13$). While \textbf{Ours} and \textbf{a+m} start at around the same L2, ours reaches a lower L2 in fewer samples.
   }
\label{fig:multi}
\end{figure}
The benefits of our model are further shown in Figure~\ref{fig:multi} which tests how well each method captures the distribution of listener-speaker dynamics in the dataset. For each method, we feed in the same 256 length test sequences, and sample $x \in [1,200]$ output listener motion sequences. From the $x$ sampled trajectories, we calculate the avg.~minimum L2 distance to ground truth. For \textbf{NN}, we sample the top 200 64-length sequences. The loss from quantizing \textbf{GT} is shown ($\text{L2}=17.13$). 
\textbf{Ours} requires fewer sampling steps to achieve motion that is closer to the actual ground truth listener motion. While \textbf{a+m} starts out at a similar L2, it takes far more samples to reach a lower L2. This indicates that our method better models the distribution of actual listener-speaker dynamics seen in the dataset.

\medskip
\noindent \textbf{TLCC Analysis.}
As mentioned in the paper, we perform a time lagged cross correlation analysis to get a sense of the leader-follower relationship between the speaker and listener, as well as to analytically find the optimal delay for our Mirror Delay baseline. We provide the full analysis for all of the baselines here in Table~\ref{tab:tlcc}. The results demonstrate that the difference between the measured TLCC between \textbf{GT} and \textbf{Ours} is not significant.

\begin{table}[t]\footnotesize
    \centering
    \begin{tabular}{@{}lcc@{}}
    \toprule
    & Expression & Rotation \\
    \midrule
    GT          & 17.3 & 16.1 \\
    NN motion   & 15.3 & 16.0 \\
    NN audio    & 17.9 & 17.5 \\
    Random      & 13.2 & 14.2 \\
    Median      & -    & - \\
    Mirror      & 0.0  & 0.0 \\
    Delayed Mirror & 17.0 & 17.0  \\
    LFI[34]     & 15.2 & 16.9  \\
    Ours        & 15.8 & 17.4 \\
    \toprule
    \end{tabular}
    \caption{Time lagged cross correlation results. We use this analysis to set the delay amount in Delayed Mirror baseline.}
    \label{tab:tlcc}
\end{table}

\medskip
\noindent \textbf{Qualitative Evaluation Details.}
For each Ours vs.~NN motion, vs.~$a+m$, vs.~GT, we generate 50 A-B tests. For each of the 50 A-B tests, we ask 3 different evaluators, totalling to 450 evaluators. Prior to the actual test, we provide a headphone check to make sure the evaluators are listening to audio. However, we do not ask additional questions that check to see if they are actually listening to the speech. The landing page describes the task and walks evaluators through 2 examples. 
To ensure the evaluators are not just randomly clicking, we include 3 questions with an obvious mismatch (one speaker laughing while the listener is neutral) twice. If the evaluator selects a different response for these duplicated questions, we do not allow them to submit. 


\section{Risks and Potential Misuse} 
While our proposed framework is intended solely for improving the naturalness of next generation human-machine interaction systems, such as virtual assistants and other entertainment applications, our AI-synthesized video output component produces highly realistic looking humans, that can be confused with a real person. Similar to many advanced video synthesis techniques (such as face-swap deepfakes, facial reenactment algorithms), such pipeline could be potentially miused by malicious actors, as the results can be highly convincing and the algorithm easy to replicate as long as the data is available. For instance, a fake conversation between two person could be constructed while it never occurred, or a fake video chat participant could be used to attend an unauthorized meeting. While it can be difficult to fully prevent the misuse of such technology, we believe that this paper can help raise awareness of such capabilities, and we advocate for a safe use of video-synthesis technique using watermarks labeling the content for example as `synthesized'. Furthermore, we believe that our speaker-to-listener translation approach and prediction models can inspire future video manipulation detection algorithms to look for additional cues in listener's performance.







\end{document}